\PassOptionsToPackage{table}{xcolor}
\documentclass[conference,a4paper]{APSIPA2025}
\usepackage{amsmath}
\usepackage{graphicx}
\usepackage{subcaption}
\usepackage{array}
\usepackage{multirow}
\usepackage{threeparttable}
\usepackage{cite}

\usepackage{geometry}

\usepackage{fancyhdr}

\usepackage{booktabs}

\usepackage{hyperref}

\usepackage[capitalize,noabbrev]{cleveref}
\usepackage{amssymb}
\usepackage[table]{xcolor}

\usepackage{setspace}
\makeatletter

\renewenvironment{thebibliography}[1]{%
  \begin{oldthebibliography}{#1}%
  \small
}{%
  \end{oldthebibliography}%
}
\makeatother

\fancypagestyle{firststyle}{
  \fancyhf{}
  \fancyhead[C]{2025 Asia Pacific Signal and Information Processing Association Annual Summit and Conference (APSIPA ASC)}
}

\begin{document}

\title{TRUST: Token‑dRiven Ultrasound Style Transfer for Cross‑Device Adaptation}

\author{
Nhat-Tuong Do-Tran\authorrefmark{1}, 
Ngoc-Hoang-Lam Le\authorrefmark{1}, 
Ian Chiu\authorrefmark{2}, 
Po-Tsun Paul Kuo\authorrefmark{3}, and 
Ching-Chun Huang\authorrefmark{1}
\\
\authorrefmark{1}National Yang Ming Chiao Tung University, Taiwan \\
\authorrefmark{2}Advantech Company, Taiwan \authorrefmark{3}Advanced Operational Development, Delta Electronics, Inc., Taiwan \\
\small \authorrefmark{1}\{tuongdotn.ee12, lengochoanglam.ee12, chingchun\}@nycu.edu.tw, \authorrefmark{2}Ian.Chiu@advantech.com.tw, \authorrefmark{3}Paul.PT.Kuo@deltaww.com
}

\maketitle
\thispagestyle{firststyle}
\pagestyle{fancy}
\thispagestyle{firststyle}
\pagestyle{fancy}

\begingroup
\hypersetup{hidelinks}
\renewcommand\thefootnote{}\footnote{Equal contribution: Nhat-Tuong Do-Tran and Ngoc-Hoang-Lam Le.}
\renewcommand\thefootnote{}\footnote{Corresponding author: Ching-Chun Huang.}
\addtocounter{footnote}{-1}
\endgroup

\vspace{-1em}
\begin{abstract}

Ultrasound images acquired from different devices exhibit diverse styles, resulting in decreased performance of downstream tasks. To mitigate the style gap, unpaired image-to-image (UI2I) translation methods aim to transfer images from a source domain, corresponding to new device acquisitions, to a target domain where a frozen task model has been trained for downstream applications. However, existing UI2I methods have not explicitly considered filtering the most relevant style features, which may result in translated images misaligned with the needs of downstream tasks. In this work, we propose TRUST, a token-driven dual-stream framework that preserves source content while transferring the common style of the target domain, ensuring that content and style remain unblended. Given multiple styles in the target domain, we introduce a Token-dRiven (TR) module that operates from two perspectives: (1) a data view--selecting ``suitable" target tokens corresponding to each source token, and (2) a model view--identifying ``optimal" target tokens for the downstream model, guided by a behavior mirror loss. Additionally, we inject auxiliary prompts into the source encoder to match content representation with downstream behavior. Experimental results on ultrasound datasets demonstrate that TRUST outperforms existing UI2I methods in both visual quality and downstream task performance.
\end{abstract}

\section{Introduction} \label{sec:intro}

Ultrasound (US) imaging is widely adopted in clinical diagnosis due to its efficiency, affordability, safety, and real-time capability \cite{UI_pros, UI_pros2}. However, images acquired from different devices or medical centers often exhibit substantial variation in visual characteristics—such as intensity, contrast, resolution, and speckle patterns—due to differences in hardware configurations and acquisition protocols. These cross-device discrepancies pose serious challenges to both automated systems and human interpretation. Deep learning models, once trained on a specific source domain, tend to suffer significant performance degradation when deployed in unseen domains. Similarly, doctors—accustomed to images from familiar devices—may face difficulty interpreting cases from new devices that display unfamiliar style distributions. While retraining on new devices is a possible solution, it is often impractical: (1) \underline{For doctors}, adapting to new device styles requires considerable effort in diagnosing new image characteristics. (2) \underline{For deployed AI models}, fine-tuning is typically not feasible due to proprietary software constraints (i.e., cannot be reconfigured model parameters). To mitigate such domain shifts, Ultrasound Style Transfer (UST) has emerged as a promising solution. Instead of retraining, UST translates the source-domain images to match the style of the target domain while preserving their structural content. In this way, both doctors and downstream models operate in a target-style environment without altering their original behavior.

\begin{figure}[!t]
    \centering
    \includegraphics[width=0.9\columnwidth]{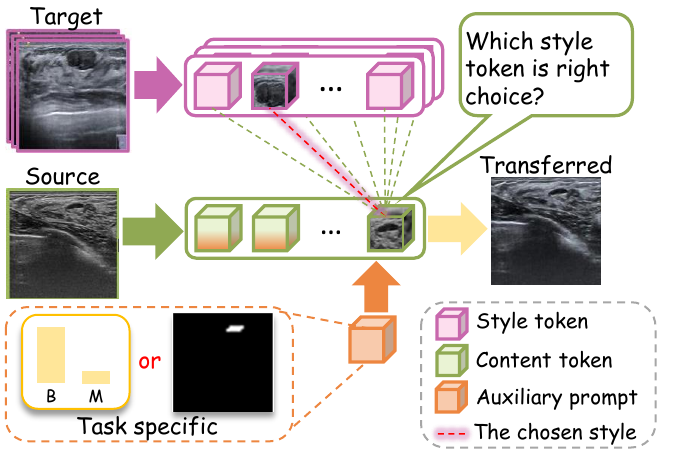}
    \caption{TRUST selects the most appropriate style token for each content token by considering both the data view (e.g., texture, pattern) and the model view (e.g., classification, segmentation), the latter enabled by auxiliary prompts injected into the content encoder.}
    \label{fig:concept}
    \vspace{-.5em}
\end{figure}

During the last decades, unpaired image-to-image translation methods have been explored, aiming to adapt images from the source domain to the target domain without requiring paired data. Among these, GAN-based methods \cite{staingan, usgan, CycleGAN} have been early solutions for bridging domain gaps, but are mainly effective for small shifts. Moreover, their shared discriminator often entangles content and style. Transformer-based models \cite{tanrscolour, S2WAT} have shown strong capabilities in disentangling content and style by separating their respective branches, effectively preventing undesired blending. Their self-attention mechanism enables a global understanding of spatial structure, which helps preserve content integrity and maintain consistent patterns across layers \cite{trans_vision, StyTr}. Despite these advantages, existing transformer-based methods often rely on pairwise style translation, which limits their ability to capture diverse styles, especially when the target domain exhibits multiple stylistic variations. Furthermore, transformer-based methods lack trustworthy token alignments, which are key to reliable downstream performance. 
 
To address the above challenges, we propose the  \textbf{TRUST}, a \textbf{T}oken-d\textbf{R}iven \textbf{U}ltrasound \textbf{S}tyle \textbf{T}ransfer for Cross-Device Adaptation. Our method begins with a disentangling strategy that separates the source and target into content and style branches, respectively. This strategy enables explicit extraction and independent processing of style and content features, allowing finer control over the transfer. Inspired by Einstein’s mindset that \textbf{“not all information is knowledge!”}, TRUST recognizes that not all style features are beneficial for transfer. Naively integrating global style tokens can lead to artifacts, such as overly bright tumor regions or suppressed tissue contrast. To this end, TRUST employs a selective mechanism that searches through a \textit{pool} of style tokens to identify those best aligned with each content token (\cref{fig:concept}). Crucially, the selection of style features is guided not only by the content tokens but also by auxiliary prompts injected into the content encoder layers. These prompts help the TR module more effectively retain task-specific and highly relevant style features. Together, the TR module and auxiliary prompts strengthen the connection between content and style branches, both at the data and model levels, enabling more accurate knowledge transfer. enabling more accurate knowledge transfer. Additionally, to align the output with the downstream task, we construct a mimic model that replicates the behavior of the black-box downstream model. Then, we use the mimic model’s predictions as supervision to align the styled image output through a behavior mirror loss. We summarize the key contributions of TRUST as follows:
\begin{enumerate}
    \item The Token-dRiven (TR) module employs attention to identify trustworthy style tokens and selectively fuses them based on content relevance, enabling more precise and natural style integration.
    \item We inject auxiliary prompts into the content encoder to enhance feature representation and more effectively minimize the domain gap.
    \item We introduce a behavior mirror loss that uses a mimic downstream to replicate black-box behavior and supervise the styled image output.
    \item Comprehensive experiments demonstrate that TRUST consistently outperforms baseline methods across multiple datasets, achieving outstanding results with well-preserved content structures and desirable style patterns.
\end{enumerate}

\begin{figure*}[!ht]
    \centering
    \includegraphics[width=0.9\textwidth]{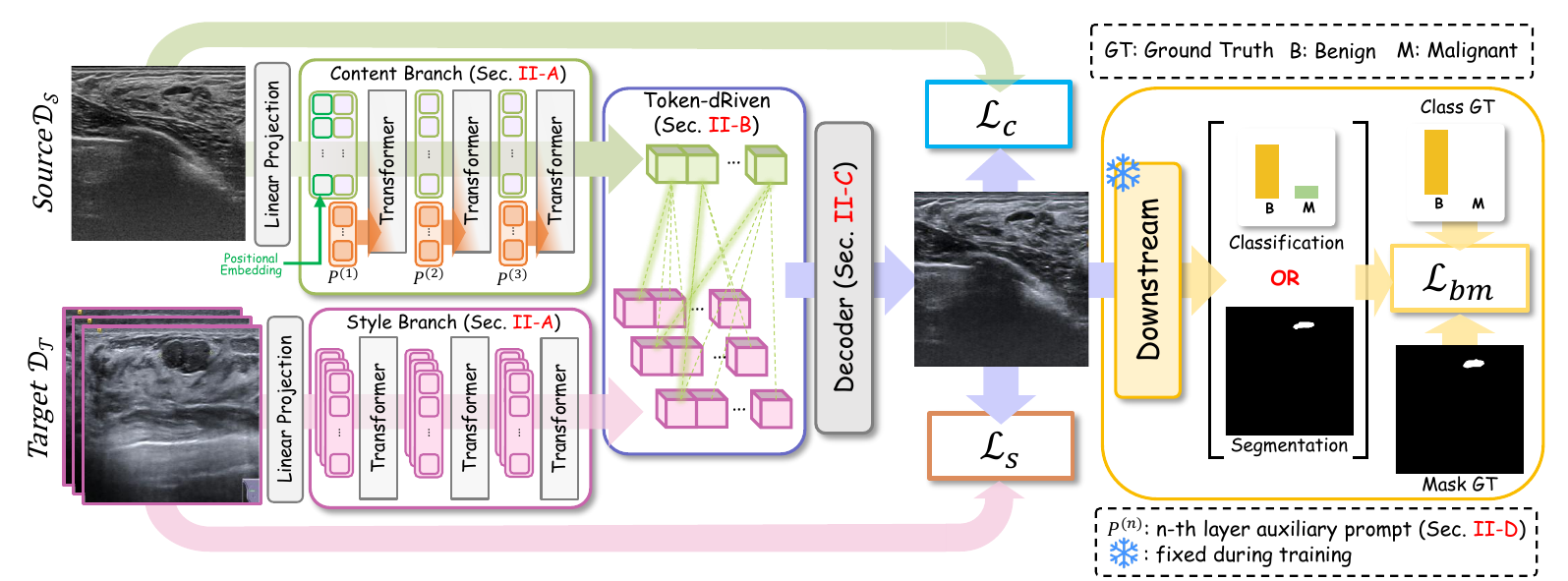}
    \vspace{-.5em}
    \caption{\textbf{Overview of the proposed TRUST framework for ultrasound style transfer}. The architecture consists of a content branch with multi-layer auxiliary prompts for data adaptation, a style branch for extracting target style features, a Token-dRiven (TR) module that fuses latent source content tokens with target style tokens to perform style transfer, and a decoder that reconstructs the stylized ultrasound images. Given a frozen downstream task network, a behavior mirror loss ($\mathcal{L}_{bm}$) is introduced to provide task-specific supervision, encouraging the source content tokens to align with downstream objectives originally tailored for the target domain.}
    \label{fig:architecture}
    \vspace{-1em}
\end{figure*}

\section{Proposed method} \label{sec:method}

\subsection{Problem definition} \label{subsec:problem}

We formulate the ultrasound style transfer for cross-device adaptation as an unpaired image-to-image translation problem.
Let $\mathcal{D}_S = \{(x_s^i, y_s^i)\}_{i=1}^{N_{labeled}} \cup \{x_s^k\}_{k=1}^{N_{unlabeled}}$ represent the source domain dataset, where a \textbf{few} $x_s^i$ denotes an labeled image with corresponding annotations $y_s^i$, while the majority of source samples $x_s^k$ are unlabeled. Similarly, $\mathcal{D}_T = \{x_t^j\}_{j=1}^{N_T}$ represents the target domain dataset without annotations, where images $x_t^j$ are acquired from different ultrasound scanners.

Our goal is to learn a translation function $G: X_S \rightarrow X_T$ capable of mapping images $x_s \in X_S$ to the target domain $\mathcal{D}_T$, thus addressing the domain shift caused by cross-device variations. The translated images $\hat{x}_t = G(x_s)$ should preserve the anatomical content structures inherent in $x_s$ while effectively adapting to the style characteristics found in $X_T$.

To achieve this, we design our style transfer network $G(x_s)$, TRUST, as illustrated in \cref{fig:architecture}. TRUST comprises three key components: \textbf{feature extractors} (\cref{subsec:feature}) to encode representations from the source and target domains, a \textbf{Token-dRiven (TR)} module (\cref{subsec:TR}) that aligns one-source tokens with multiple-target tokens via data-view and model-view, and a \textbf{decoder} (\cref{subsec:decoder}) that reconstructs stylized images while preserving source content and transferring target style. In addition, \textbf{auxiliary prompts} (\cref{subsec:prompts}) are injected into the content encoder to enhance feature representations.

\subsection{Feature extractors} \label{subsec:feature}

Our feature extraction module consists of two Transformer-based branches: a \textbf{content branch} and a \textbf{style branch}. Both branches are constructed with multiple Transformer layers \cite{attention}, leveraging the self-attention mechanism to effectively preserve structural information and progressively enhance semantic representations across layers.

\textbf{Content branch.} Focuses on capturing structural and anatomical details from the source domain. Transformer-based architectures are particularly suitable for this purpose because they keep resolution unchanged, while enriching the semantic information layer by layer, as demonstrated in \cite{StyTr}. Given a source image $x_s \in \mathbb{R}^{H \times W \times C}$, the image is first divided into non-overlapping patches of size $P \times P$, resulting in $N = \frac{H \cdot W}{P^2}$ patches. After linear projection, the patch tokens are added with \textit{learnable positional embeddings} to encode spatial information. The source tokens are then processed by the transformer layers as follows:
\begin{equation}
    F_s = E_s(x_s, P), \quad F_s \in \mathbb{R}^{N \times d},
\end{equation}
where $E_s$ represents the source encoder, $N$ is the number of patches, and $d$ is the embedding dimension of each token. $P$ is a set of auxiliary prompts that can be inserted into the token sequence, as described in \cref{subsec:prompts}.

\textbf{Style branch.} Processes target domain images to extract target tokens that reflect the distributional and stylistic patterns of the target domain. Given a batch of target images $X_T^{batch} = \{x_t^1, x_t^2, \dots, x_t^B\}$, where each $x_t^j \in \mathbb{R}^{H \times W \times C}$ and $B$ is the batch size, the encoder produces target tokens:
\begin{equation}
    F_t = E_t(X_T^{batch}), \quad F_t \in \mathbb{R}^{B \times N \times d},
\end{equation}
where $E_t$ denotes the target encoder.

The use of stacked Transformer layers in both branches ensures that spatial information is preserved while semantic understanding is progressively improved, enabling effective feature extraction for both content and style domains \cite{StyTr}.

\subsection{Token-dRiven module} \label{subsec:TR}

To effectively select suitable and informative multiple-target tokens for one-source tokens, we propose the TR module via data-view and model-view.

\textbf{Data-view.}
From a data perspective, we design the TR module to align each source token with target tokens from multiple samples by computing a correlation matrix that captures their pairwise relationships.
Formally, given source tokens $F_s \in \mathbb{R}^{N \times d}$ and target tokens $F_t \in \mathbb{R}^{B \times N \times d}$, the correlation matrix $M \in \mathbb{R}^{N \times (B \cdot N)}$ is computed as:
\begin{equation}
    M_{ij} = \text{dot}(F_s^i, F_t^j),
\end{equation}
where $\text{dot}(\cdot,\cdot)$ represents the dot product between each source token and the aggregated target tokens across the batch.

The aligned features $F_{align} \in \mathbb{R}^{N \times d}$ are then obtained by aggregating the target tokens weighted by this correlation matrix, followed by a skip connection to preserve the original source content:
\begin{equation}
    F_{align} = F_s + M F_t.
\end{equation}
In this way, each source token selectively attends to the most relevant target tokens across different samples, enabling style adaptation while maintaining source content structure.

\textbf{Model-view.}
To further enforce consistency with downstream tasks, we introduce the \textbf{behavior mirror loss} ($\mathcal{L}_{bm}$), which leverages supervision signals from a downstream model trained on the labeled source samples, $D_s^{labeled}=\{(x_s^i, y_s^i)\}_{i=1}^{N_{labeled}}$. The behavior mirror loss is formulated as:
\begin{equation}
    \mathcal{L}_{bm}\big(G(x_s^i), y_s^i\big) = \ell_{down}\big(f_{downs}(G(x_s^i)), y_s^i\big),
\end{equation}
where $f_{downs}(\cdot)$ is the downstream model, selected according to the deployment scenario (i.e., ViT-B/16 \cite{ViT} for classification and SAMUS \cite{SAMUS} for segmentation), and $\ell_{down}(\cdot, \cdot)$ represents the corresponding loss function. By minimizing $\mathcal{L}_{bm}$, the source tokens $F_s$ are directly optimized toward the downstream objective, ensuring that the translated image $G(x_s^i)$ produces predictions consistent with the label $y_s^i$.

\subsection{Decoder} \label{subsec:decoder}
The decoder reconstructs the stylized image from the aligned token features through a convolutional architecture following previous style transfer works \cite{StyTr, SANet}. Given the aligned tokens $ F_{align} \in \mathbb{R}^{N \times d} $, the decoder first reshapes them into a spatial feature map of size $ H' \times W' \times d $, where $ H' $ and $ W' $ equal to $\sqrt{N}$. The final output is obtained via CNN-based decoder layers as follows:

\begin{equation}
    \hat{x}_t = D(F_{align}),
\end{equation}
where $ D $ denotes the decoder, and $ \hat{x}_t \in \mathbb{R}^{H \times W \times 3} $ is the reconstructed stylized image.

\subsection{Auxiliary prompts} \label{subsec:prompts}

To enhance content representations for better compatibility with the downstream model. We inspired of VPT \cite{VPT} to inject a set of learnable prompt tokens $P=\{P^{(l)} \in \mathbb{R}^{L_p \times d} \}_{l=1}^{L}$ into the source encoder (\cref{subsec:feature}), where $L$ is the number of transformer layers, $L_p$ is the number of prompt tokens per layer, and $d$ is the token dimension.

At each transformer layer, a unique set of prompt tokens $P^{(l)}$ is prepended to the input token sequence. The prompt-enhanced content extraction is thus formulated as:
\begin{equation}
    F_s = E_s\big(\{P^{(l)} \oplus x_s^{(l)}\}_{l=1}^{L}\big),
\end{equation}
where $x_s^{(l)}$ denotes the input tokens at layer $l$, and $\oplus$ represents the concatenation operation at the token level.

\begin{table*}[!t]
\centering
\caption{
Performance comparison of style transfer methods on $6$ cross-device transfer tasks for both classification and segmentation. Classification performance is evaluated by Accuracy (Acc) and Area Under the Curve (AUC), while segmentation performance is evaluated by Dice score (Dice) and Intersection over Union (IoU). 
The results are reported in the order: Acc / AUC / Dice / IoU. \colorbox{gray!20}{w/o ST} indicates the baseline setting \textit{without Style Transfer}. The top and second-best results are highlighted in \textbf{bold} and \underline{underline}.
}
\vspace{-.5em}
\setlength{\tabcolsep}{3pt}
\resizebox{\textwidth}{!}{
\begin{tabular}{l|c|c|c|c|c|c}
\toprule
\textbf{Method} 
& {UCLM $\rightarrow$ BUSI} 
& {BUSI $\rightarrow$ UCLM} 
& {UCLM $\rightarrow$ UDIAT} 
& {UDIAT $\rightarrow$ UCLM} 
& {BUSI $\rightarrow$ UDIAT} 
& {UDIAT $\rightarrow$ BUSI} \\
\midrule
\rowcolor{gray!20} {w/o ST} & 70.0/74.8/\underline{77.1}/\underline{66.5} & 65.6/68.8/\underline{77.7}/\underline{68.0} & 63.8/68.2/82.2/72.1 & 75.5/77.1/85.1/75.6 & \underline{73.3}/73.2/79.5/70.6 & 85.7/\underline{91.9}/\underline{84.3}/\underline{74.8} \\
CycleGAN \cite{CycleGAN} & 51.3/64.6/65.2/52.4 & \underline{69.2}/72.4/75.7/65.5 & 56.3/56.9/74.7/63.9 & 67.4/69.2/80.4/69.2 & 70.8/72.0/79.7/70.4 & 79.5/81.6/82.5/72.6 \\
DiscoGAN \cite{DiscoGAN} &70.0/72.0/67.9/56.3 & 63.1/\underline{74.1}/73.2/63.2 & \underline{65.0}/\underline{70.1}/42.6/30.6 & 69.4/\underline{81.4}/44.1/33.0 & 72.3/\underline{73.8}/77.4/67.3 & \underline{87.8}/88.0/79.2/68.0 \\
S2WAT \cite{S2WAT} & \underline{72.5}/75.5/76.2/65.9 & 62.6/57.4/77.4/67.5 & \underline{65.0}/45.7/80.5/70.0 & 75.5/71.6/84.6/75.2 & 71.8/73.4/80.1/71.0 & 85.7/93.4/81.8/72.1 \\
TransColors \cite{tanrscolour} & \underline{72.5}/\underline{77.3}/75.9/65.4 & 64.1/65.0/\underline{77.7}/67.9 & 62.5/68.4/\underline{82.6}/\underline{72.6} & \underline{77.6}/73.9/\underline{85.7}/\underline{76.2} & 69.2/71.1/\underline{80.4}/\underline{71.4} & 83.7/92.1/83.4/74.0 \\

\textbf{TRUST} & \textbf{80.0}/\textbf{81.9}/\textbf{79.9}/\textbf{69.7} & \textbf{71.8}/\textbf{77.9}/\textbf{78.6}/\textbf{69.0} & \textbf{72.5}/\textbf{76.7}/\textbf{83.3}/\textbf{73.4} & \textbf{85.7}/\textbf{89.7}/\textbf{86.9}/\textbf{77.9} & \textbf{79.5}/\textbf{84.5}/\textbf{80.7}/\textbf{71.8} & \textbf{89.8}/\textbf{94.9}/\textbf{85.4}/\textbf{76.5} \\
\bottomrule
\end{tabular}
}
\label{tab:transfer_performance}
\end{table*}

\begin{figure*}[!t]
    \centering
    \includegraphics[width=\textwidth]{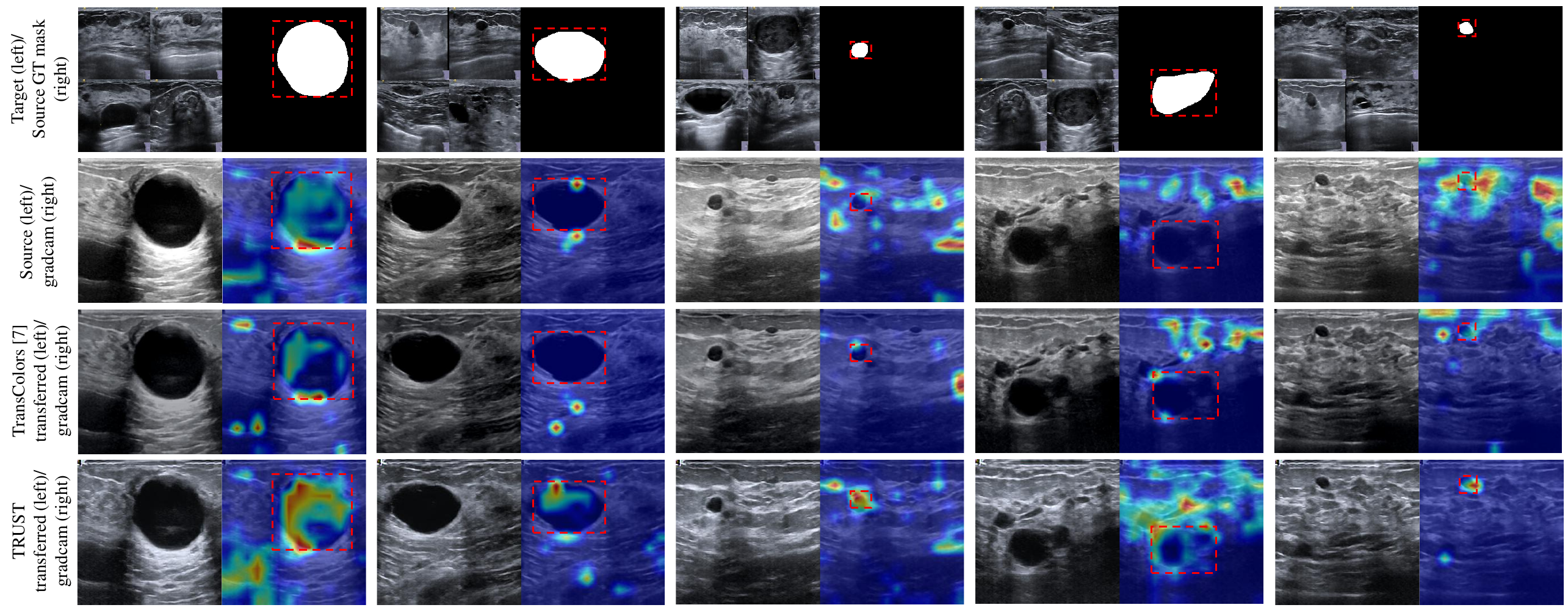}
    \caption{\textbf{GradCAM \cite{gradcam} comparison of style transfer methods}, including Source without transfer (Source), TransColors~\cite{tanrscolour}, and our proposed TRUST. The GradCAM visualizations highlight the attention regions of a frozen downstream classifier on the BUSI$\rightarrow$UCLM transfer task. The source mask (top-right) delineates the ground-truth tumor region that the model should attend to for accurate classification, while the target images (top-left) serve as style references for the style transfer methods. Notably, the transferred source images (bottom-left) produced by TRUST exhibit more concentrated attention within the tumor region, thereby enhancing downstream classification performance.}

    \label{fig:qualitative}
\end{figure*}

\subsection{Training Scheme} \label{subsec:training}
Following previous style transfer works \cite{SANet, StyTr, S2WAT}, we adopt a content loss ($\mathcal{L}_{c}$) to preserve source structure and a style loss ($\mathcal{L}_{s}$) to match target style statistics, both computed from the feature representations of a pre-trained VGG network. The objective function to optimize TRUST is defined as:
\begin{equation}
    \mathcal{L}_{total} = \mathcal{L}_{c}(\hat{x}_t, x_s) + \mathcal{L}_{s}(\hat{x}_t, x_t) + \mathcal{L}_{bm}(\hat{x}_t, y^i_s).
\end{equation}

\section{Experimental results} \label{sec:experiments}
\subsection{Experimental setup}
\textbf{Dataset.} 
We conduct experiments on three publicly available breast ultrasound datasets: BUSI \cite{BUSI}, UDIAT \cite{UDIAT}, and UCLM \cite{BUS-UCLM}. Each dataset is randomly divided into training and testing subsets with a ratio of $7{:}3$. In each experimental setting, one dataset (e.g., BUSI) is assigned as the \textit{source domain}, $\mathcal{D}_S$, while another serves as the \textit{target domain}, $\mathcal{D}_T$ (e.g., UDIAT or UCLM). The proposed TRUST framework is trained on the source training set with style information transferred from the target training set, and evaluated on the source testing set. The testing set of the target domain is reserved for selecting the \underline{best downstream model} during validation. Notably, the target labels are only used to train the downstream model.

\textbf{Implementation details.}
All experiments are conducted on a single NVIDIA RTX $4090$ GPU with input images resized to $256 \times 256$. In the labeled source set $D_s^{labeled}=\{(x_s^i, y_s^i)\}_{i=1}^{N_{labeled}}$, only $20$ samples per class are available. The entire TRUST framework is initialized using Xavier uniform initialization \cite{glorot2010understanding}. Both the source encoder $E_s$ and target encoder $E_t$ consist of $3$ Transformer \cite{attention} layers with $L_p = 1024$ prompt tokens per layer and a patch size of $P=8$. We adopt the Adam optimizer with an initial learning rate of $5 \times 10^{-4}$, following the warm-up adjustment strategy \cite{warmlr}, and use a batch size of $B=6$. The total number of iterations is set to $20,000$. The \underline{classification downstream} is ViT-B/16 \cite{ViT}, trained using SGD with a learning rate of $0.001$, momentum of $0.9$, weight decay of $0.0005$, and batch size of $16$. For the \underline{segmentation downstream}, we employ SAMUS \cite{SAMUS} with its default training configuration.

%
\begin{table}[!t]
\centering
\caption{
\textbf{Ablation study on different components of our framework}. “w/o ST” denotes the case without style transfer. CA refers to the baseline model employing Cross Attention, TR represents the Token-dRiven module, and TRUST integrates TR with auxiliary prompts for enhanced style adaptation.
}
\vspace{-.5em}
\setlength{\tabcolsep}{3pt}
\resizebox{\columnwidth}{!}{
\begin{tabular}{l|c|c|c}
\toprule
\textbf{Settings} 
& {BUSI $\rightarrow$ UCLM} 
& {UCLM $\rightarrow$ UDIAT} 
& {UDIAT $\rightarrow$ BUSI} \\
\midrule
\rowcolor{gray!20} w/o ST & 65.6/68.8/77.7/68.0 & 63.8/68.2/82.2/72.1 & 85.7/91.9/84.3/74.8 \\
CA & 66.7/64.9/77.8/68.0 & 60.0/66.3/82.9/72.9 & 87.8/92.3/83.6/73.9 \\
TR & 68.7/76.7/78.5/68.6 & 71.3/76.2/83.1/72.9 & 89.8/92.5/85.4/76.2 \\
\textbf{TRUST} & 71.8/77.9/78.6/69.0 & 72.5/76.7/83.3/73.4 & 89.8/94.9/85.4/76.5 \\
\bottomrule
\end{tabular}
}
\label{tab:ablation_study}
\vspace{-.2em}
\end{table}

\subsection{Comparison results}
In this section, we provide quantitative and qualitative comparisons between the proposed \textbf{TRUST} and prior style transfer methods, including GAN-based approaches (CycleGAN~\cite{CycleGAN}, DiscoGAN~\cite{DiscoGAN}) and Transformer-based approaches (S2WAT~\cite{S2WAT}, TransColors~\cite{tanrscolour}). These comparisons demonstrate that TRUST effectively preserves the structural content of the source and injects suitable, task-optimal styles from the target domain. Surprisingly, TRUST achieves state-of-the-art results on $6$ cross-device ultrasound tasks.

\textbf{Quantitative results.}
Table~\ref{tab:transfer_performance} reports the classification (Accuracy/AUC) and segmentation (Dice/IoU) performance before and after applying style transfer. In particular, TRUST substantially shows improvements compared to without style transfer (w/o ST) and GAN-based works. Furthermore, compared to the Transformer-based method TransColors, TRUST achieves notable gains of $+10.0\%$ in Accuracy and $+8.3\%$ in AUC on the UCLM $\rightarrow$ UDIAT classification task. In segmentation, it outperforms S2WAT under the UDIAT $\rightarrow$ BUSI setting, with improvements of $+3.6\%$ in Dice and $+4.4\%$ in IoU. These results indicate that S2WAT or TransColors relies heavily on one-to-one (source-target image) pairings, limiting their ability to capture the common target styles. Moreover, their naive style injection often introduces unwanted noise into the source content, resulting in degraded performance—even below w/o ST—on tasks, such as BUSI $\rightarrow$ UCLM and UDIAT $\rightarrow$ BUSI.

\textbf{Qualitative results.} In \cref{fig:qualitative}, we present visual comparisons and Grad-CAM \cite{gradcam} results on the test set, revealing several key observations. First, TRUST outperforms TransColors in capturing diverse styles within a specific target domain. Its outputs clearly reflect this strength, showing the closest resemblance to the target samples in color and brightness. Second, TRUST generates finer tissue details and clearer tumor boundaries, highlighting the advantage of using reliable information. Third, TRUST improves semantic cues for downstream classification, as shown by Grad-CAM maps that focus more precisely on tumors and their boundaries, indicating more effective attention to diagnostically relevant regions. Overall, TRUST achieves precise style transfer and task-focused attention, providing clear advantages over conventional approaches that treat all target information equally.

\begin{figure}[h]
\centering
\begin{subfigure}[t]{0.48\columnwidth}
\centering
\includegraphics[width=\linewidth]{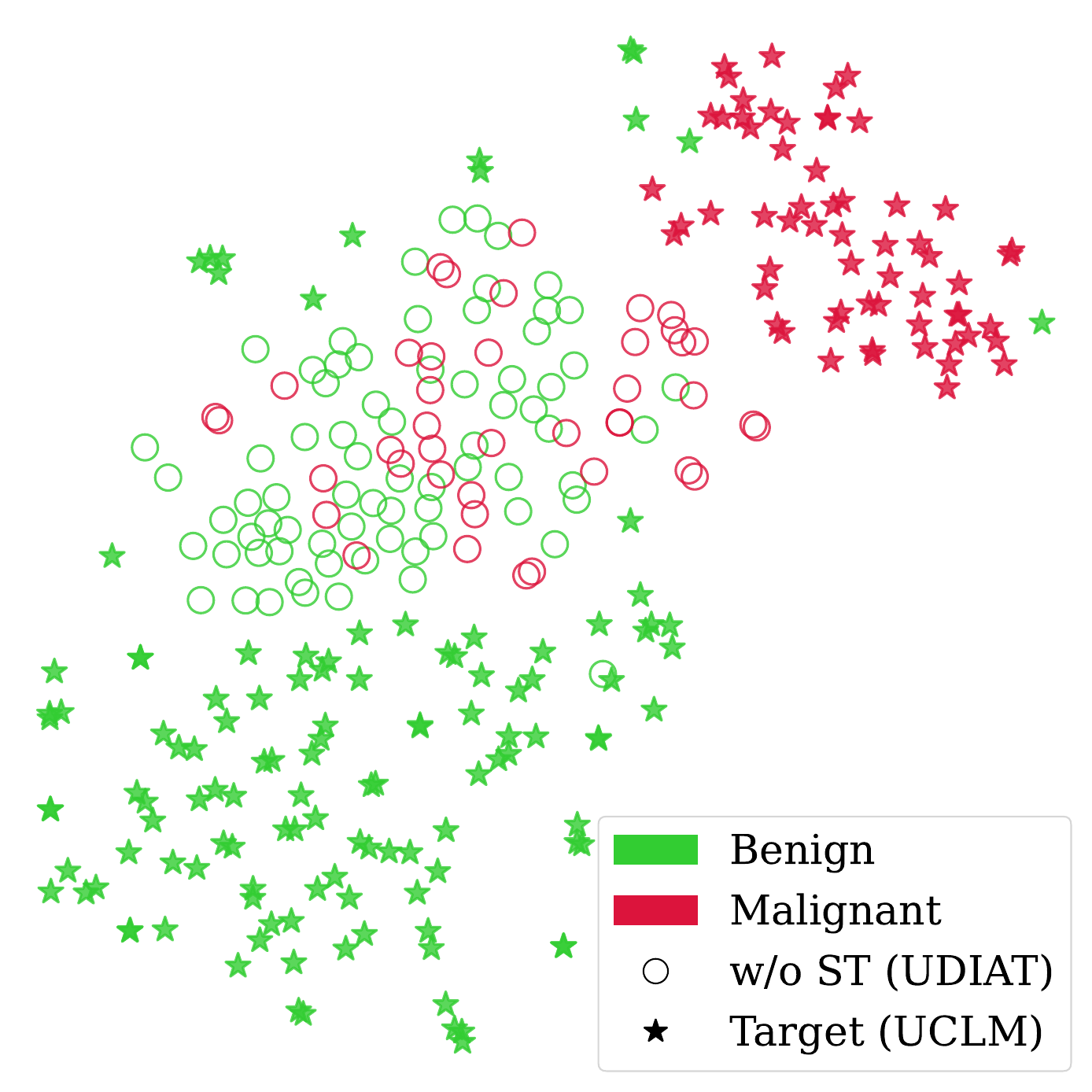}
\caption{Before Style Transfer}
\label{fig:tsne_a}
\end{subfigure}%
\hfill
\begin{subfigure}[t]{0.48\columnwidth}
\centering
\includegraphics[width=\linewidth]{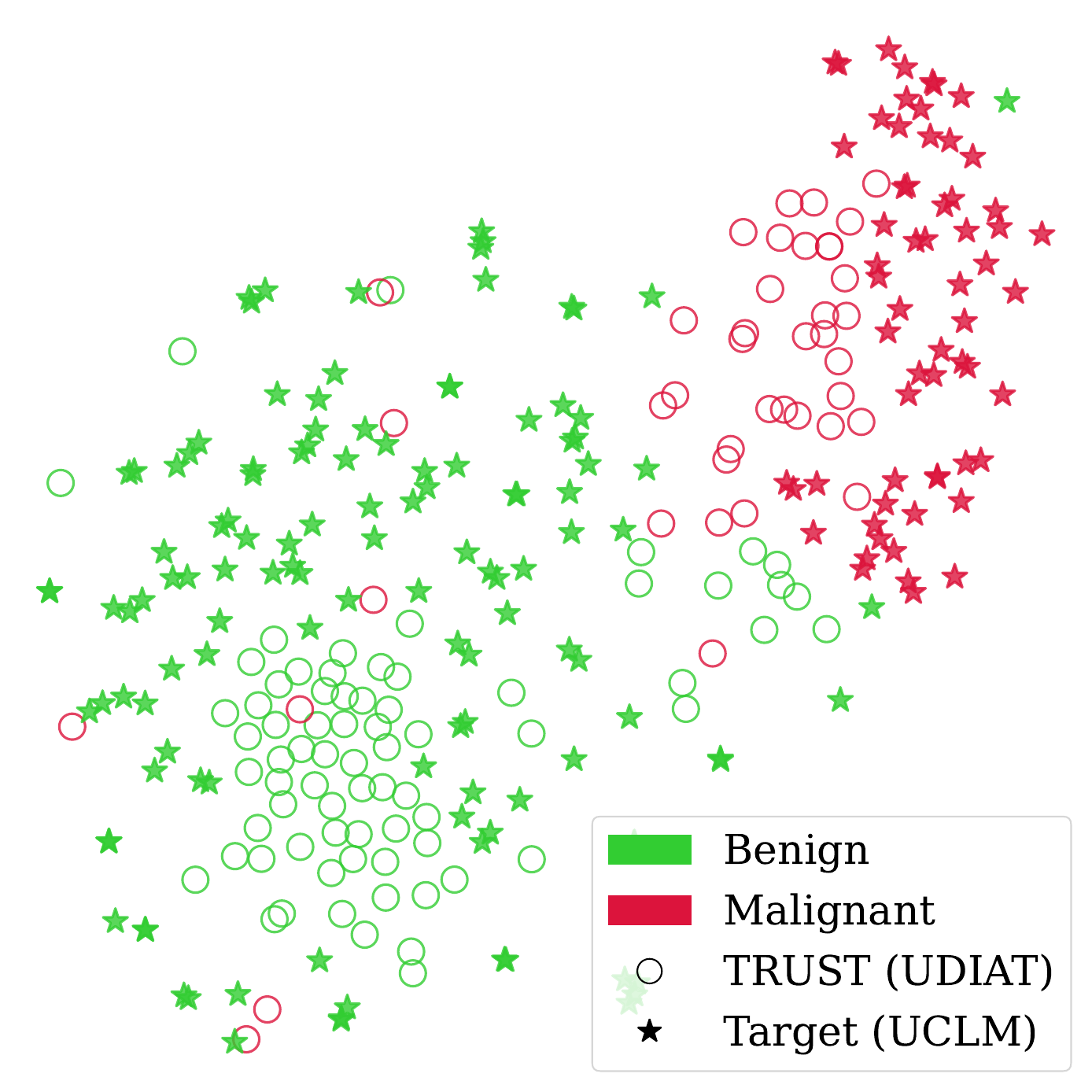}
\caption{After Style Transfer}
\label{fig:tsne_b}
\end{subfigure}
\caption{t-SNE~\cite{tsne} visualization of feature distributions from the downstream classifier on the UDIAT $\rightarrow$ UCLM task. Each point represents a sample, where \textcolor{green}{green} and \textcolor{red}{red} denote benign and malignant classes, respectively. $\star$ marks target samples (UCLM), while $\circ$ denotes source samples (UDIAT), either original (w/o ST) or translated (TRUST).}
\label{fig:style_transfer}
\vspace{-1em}
\end{figure}

\section{Analysis} \label{sec:analysis}

\textbf{Ablation studies.}
Table~\ref{tab:ablation_study} presents an ablation study across three cross-device settings to evaluate the effectiveness of each component in TRUST. Applying naive cross attention (CA) introduces noisy target tokens and can degrade performance—for example, on UCLM $\rightarrow$ UDIAT, classification accuracy drops from $63.8$ (w/o ST) to $60.0$. This supports our key motivation: \textit{“Not all information is knowledge.”} 
Replacing CA with our Token-dRiven (TR) module leads to consistent improvements across all tasks. Compared to w/o ST, TR improves classification accuracy by $+3.1\%$, $+7.5\%$, and $+4.1\%$ for BUSI $\rightarrow$ UCLM, UCLM $\rightarrow$ UDIAT, and UDIAT $\rightarrow$ BUSI, respectively. Similar trends are observed in segmentation, where TR enhances Dice scores by up to $+0.8\%$ and IoU by up to $+0.6\%$ on the BUSI $\rightarrow$ UCLM task.
Finally, integrating TR with auxiliary prompts further refines content representations. TRUST achieves the best overall performance, notably improving AUC from $92.5$ to $94.9$ and IoU from $76.2$ to $76.5$ on UDIAT $\rightarrow$ BUSI. These results confirm the complementary benefits of prompt tuning in guiding source tokens toward downstream predictions.

\textbf{Feature space.}
\cref{fig:style_transfer} presents t-SNE~\cite{tsne} visualizations of feature distributions from the downstream classifier (ViT-B/16) under the UDIAT $\rightarrow$ UCLM task. Before style transfer (\cref{fig:tsne_a}), a clear domain gap exists between source samples ($\circ$) and target samples ($\star$), leading to potential misclassification in the absence of style adaptation (w/o ST). After applying TRUST (\cref{fig:tsne_b}), the translated source features are better aligned with the target distribution. Through visualizing, we can conclude that TRUST facilitates domain-invariant feature learning while preserving class-discriminative structures essential for downstream tasks.

\noindent \textbf{Computational cost.} We further analyze the efficiency of TRUST in terms of floating-point operations (FLOPs). The overall complexity of TRUST amounts to $138.63$G FLOPs for an input resolution of $256 \times 256$.

\section{Conclusion} \label{sec:conlusion}
In this work, we present \textbf{TRUST}, a novel token-driven framework for structure-preserving style transfer in cross-device ultrasound analysis. TRUST introduces a content-aware mechanism that selectively integrates \underline{target style} cues \textbf{without distorting} the \underline{anatomical structure} of the source. Through extensive experiments on $6$ cross-domain tasks, TRUST consistently outperforms existing GAN-based and Transformer-based methods across both classification and segmentation, achieving state-of-the-art performance. Furthermore, ablation studies and t-SNE visualizations confirm that TRUST produces domain-invariant and class-discriminative features. These results demonstrate the potential of TRUST as a reliable solution for real-world deployment across heterogeneous ultrasound devices.

\noindent \textbf{Acknowledgement.} This work was financially supported in part (project number: 112UA10019) by the Co-creation Platform of the Industry Academia Innovation School, NYCU, under the framework of the National Key Fields Industry-University Cooperation and Skilled Personnel Training Act, from the Ministry of Education (MOE) and industry partners in Taiwan.  It also supported in part by the National Science and Technology Council, Taiwan, under Grant NSTC-114-2218-E-A49 -024, - Grant NSTC-112-2221-E-A49-089-MY3, Grant NSTC-114-2425-H-A49-001, Grant NSTC-113-2634-F-A49-007, Grant NSTC-112-2221-E-A49-092-MY3, and in part by the Higher Education Sprout Project of the National Yang Ming Chiao Tung University and the Ministry of Education (MOE), Taiwan. It is also partly supported by MediaTek Inc., Hon Hai Research Institute, and Industrial Technology Research Institute.

\bibliographystyle{IEEEtran}
\bibliography{mybib}


\end{document}